\documentclass[11pt]{article}
\usepackage{enumitem}

\usepackage[final]{acl}
\usepackage{times}
\usepackage{latexsym}
\usepackage{booktabs}
\usepackage{multirow}
\usepackage{array}
\usepackage{graphicx}
\usepackage{caption}
\usepackage[ruled,vlined]{algorithm2e}
\usepackage{amsmath}
\usepackage{amsfonts}
\usepackage{amssymb}
\usepackage[T1]{fontenc}

\usepackage[utf8]{inputenc}
\usepackage{pgfplots}
\pgfplotsset{compat=1.18} 

\usepackage{microtype}

\usepackage{booktabs}

\usepackage{inconsolata}

\title{Causal-Audit: Explicit and Auditable Graph-based Reasoning via Target-Aware Causal Chain Construction}

\author{Su Lan, Xuefei Yin, Yanming Zhu, \and Alan Wee-Chung Liew \\
        School of Information and Communication Technology, Griffith University, QLD, AU}

\begin{document}
\maketitle
\begin{abstract}

Causal and intervention-based question answering is fundamental to advancing large language models (LLMs) toward reasoning beyond surface-level correlations and understanding underlying causal mechanisms. However, existing LLM-based methods often rely on implicit language-level reasoning, resulting in opaque causal assumptions, unverifiable reasoning paths, and fragile predictions under complex interventions, particularly in context-free settings. In this paper, we propose an explicit and auditable causal reasoning framework for context-free intervention-based question answering. Our method formulates causal inference as structured reasoning over an explicit causal graph through four modular stages, rather than implicit end-to-end prediction. A key innovation is a target-aware causal graph construction strategy that treats the target variable as a core constraint during graph expansion, effectively suppressing irrelevant variables, spurious causal relations, and reasoning noise. We further introduce a path-level causal evidence aggregation mechanism that combines multiple causal paths while modeling both reinforcing and counteracting effects, enabling robust decision-making beyond single-chain reasoning. Extensive experiments on three benchmarks demonstrate that our framework consistently outperforms existing LLM-based methods while providing interpretable and auditable causal reasoning traces.

\end{abstract}

\section{Introduction}
\label{sec:Introduction}

Causal and intervention-based question answering is widely regarded as a critical step in advancing large language models (LLMs) from powerful language processors toward reliable reasoning systems \cite{wu2024causality}. Such questions ask how outcomes would change under hypothetical interventions rather than predicting what is typically observed, requiring models to reason about causal effects instead of exploiting statistical regularities~\cite{jin2023cladder,zhou2024causalbench}. Formally, it concerns the effect of actively manipulating a variable on a target outcome, often involving multiple interacting factors and indirect causal pathways. 

Among different settings, \emph{context-free} causal question answering, where models must infer causal effects without access to supporting passages or explicit background context, is particularly challenging~\cite{pearl2018why}. In this setting, answers cannot be derived via evidence extraction or surface-level alignment, but instead require the internal construction and evaluation of plausible causal mechanisms. In this work, we focus on this challenging yet fundamental setting of \emph{context-free causal question answering}. 

Recent advances in LLMs have led to encouraging performance gains on causal and counterfactual benchmarks, largely driven by end-to-end prediction or Chain-of-Thought style prompting~\cite{wei2022cot,kojima2022zeroshotcot}. However, a growing body of empirical studies suggests that such improvements often arise from pattern matching or shallow heuristics rather than genuine causal understanding~\cite{miller2025counterfactual}.
LLMs have been shown to struggle with intervention-level reasoning, multi-variable causal dependencies, and generalization beyond observed correlations, especially in settings that require reasoning about unseen or counterfactual scenarios~\cite{jin2023cladder,wang-2024-causalbench,zhou2024causalbench}.

Several methods have been proposed to enhance LLMs for causal question answering, but most perform reasoning primarily at the language level, implicitly delegating causal inference to token- or semantic-level generation~\cite{wei2022cot}. Paradigms such as Chain-of-Thought (CoT)~\cite{wei2022chain}, Tree-of-Thought~\cite{yao2023tree}, and Graph-of-Thought~\cite{besta2024graphofthoughts} organize textual rationales without explicit variable-level causal structure or intervention semantics, making them vulnerable to spurious yet semantically related information. Although some methods introduce causal supervision during training~\cite{li2025mitigating}, inference typically collapses to direct question answering without explicit construction or auditing of causal graphs, leaving causal pathways, effect propagation, and interference from spurious variables unobservable and unverifiable~\cite{gendron2024counterfactual}. Recent studies have consequently raised concerns about the reliability and faithfulness of such implicit reasoning, showing that generated rationales may not causally support final predictions~\cite{paul-etal-2024-making,chu2025towards}. Moreover, methods that explicitly construct graphs from textual evidence rely on context-rich inputs and often fail in context-free settings, where relevant variables and relations cannot be reliably grounded~\cite{tandon2019wiqa}. 

To address these limitations, we propose an explicit and auditable causal reasoning framework for intervention-based question answering. Rather than relying on implicit end-to-end prediction or free-form rationales, we formulate causal reasoning as structured inference over an explicit causal graph, exposing intermediate causal hypotheses and enabling inspection of the reasoning process via four modular stages. A key contribution is a \emph{target-aware} causal graph construction strategy that treats the target variable as a core constraint during graph expansion, effectively suppressing irrelevant variables, spurious relations, and reasoning noise. Moreover, we introduce a path-level causal evidence aggregation mechanism that combines multiple causal paths and models both reinforcing and counteracting effects, moving beyond single-chain reasoning. These designs transform LLMs from implicit end-to-end ``reasoners" into constrained ``causal evaluators", offering a new paradigm that integrates LLMs with symbolic causal reasoning for reliable intervention-based question answering. Extensive experiments on three datasets demonstrate consistent performance gains over existing LLM-based methods while providing interpretable and auditable reasoning traces.

The main contributions are summarized below:
\begin{itemize}
\item We formulate context-free intervention-based question answering as an explicit and auditable causal reasoning problem, moving beyond implicit language-level inference and providing a structured framework for reliable intervention reasoning with LLMs.

\item We propose a target-aware causal graph construction that explicitly treats the target variable as a core constraint during graph expansion, effectively suppressing irrelevant variables, spurious causal relations, and reasoning noise in traditional graph-based reasoning. 
\item We introduce a path-level causal evidence aggregation mechanism that validates and combines multiple causal paths while modeling both reinforcing and counteracting effects, enabling robust decision making beyond single-chain or end-to-end methods.

\item Extensive experiments on three benchmarks demonstrate that our framework consistently outperforms existing LLM-based reasoning methods while providing interpretable and auditable causal reasoning traces.

\end{itemize}

\section{Related Work}
\subsection{Direct LLM}
Recent benchmarks such as CausalBench~\cite{wang-2024-causalbench,zhou2024causalbench}, CLadder~\cite{jin2023cladder}, and CausalProbe~\cite{chi2024causalprobe} show that, despite strong surface-level performance, LLMs often rely on shallow correlations rather than genuine causal understanding, especially under complex interventions. Counterfactual and interventional reasoning further emerges inconsistently through in-context learning and remains fragile to prompt design and distribution shifts~\cite{miller2025counterfactual}. These reveal a persistent gap between correlation-based inference and causal structure learning.

\begin{figure*}[t]
    \centering
    \includegraphics[width=\linewidth]{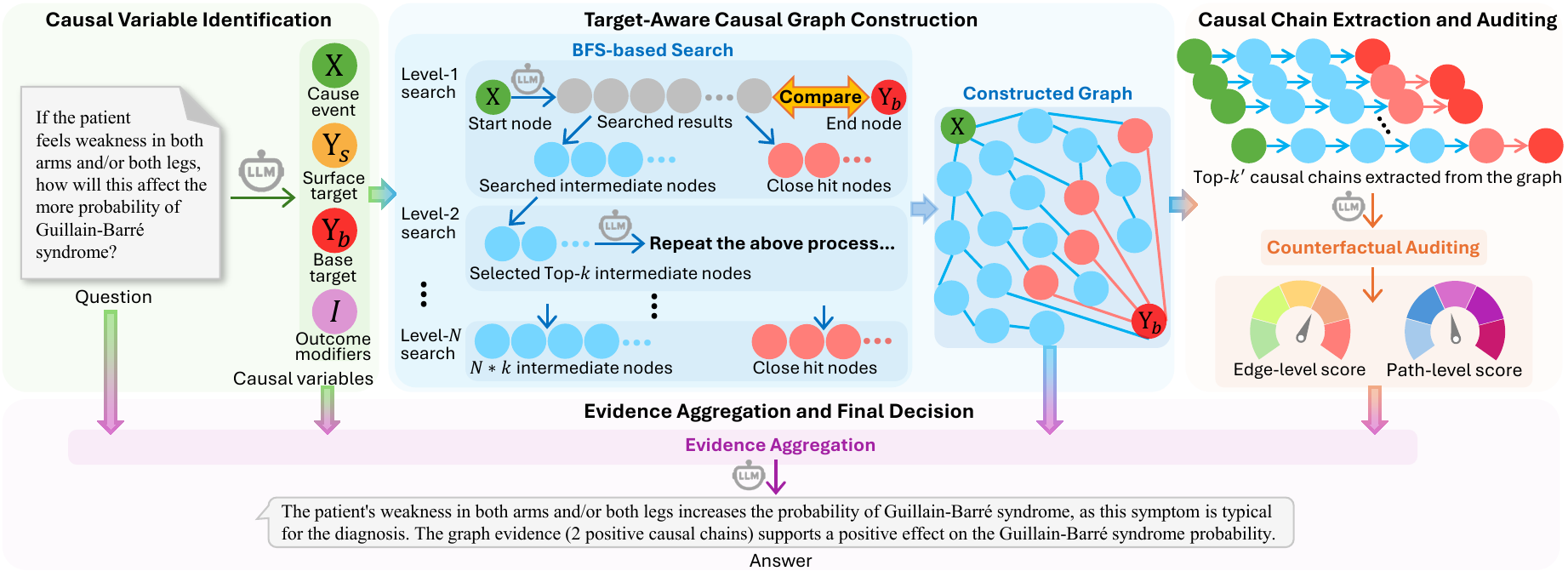}
    \caption{Overview of the proposed explicit causal reasoning framework for context-free intervention-based question answering. The framework extracts structured causal variables from the question, constructs a target-aware causal graph via iterative search, audits candidate causal chains through counterfactual validation, and aggregates evidence from multiple stages to produce an interpretable intervention decision. Detailed algorithmic descriptions are provided in Appendix~\ref{app:algo}.}
    \label{fig:pipeline}
\end{figure*}

\subsection{Chain-of-Thought Reasoning}
CoT prompting improves multi-step reasoning~\cite{wei2022cot,kojima2022zeroshotcot}, with self-consistency further enhancing robustness~\cite{wang2022selfconsistency}. However, CoT-based methods rely on unstructured natural language rationales, lack causal and structural constraints, and are often not causally faithful to final predictions~\cite{paul-etal-2024-making,chu2025towards}. Graph-based prompting and neuro-symbolic methods introduce structural scaffolding~\cite{besta2024graphofthoughts,FangD0SCPW24} but do not explicitly encode causal semantics. In contrast, our work grounds graph construction in causal relations and validates reasoning at the path level. 

\subsection{Causal Graph-based Reasoning}
Integrating causal graphs and causal inference into LLM reasoning has emerged as a promising direction for improving faithfulness and interpretability. Existing methods incorporate causal structures~\cite{wang2024core} or employ causal objectives and structured supervision~\cite{dong2025care,li2025mitigating}, yet intervention-based analyses reveal frequent violations of causal faithfulness~\cite{paul-etal-2024-making}. Moreover, most methods rely on predefined or externally provided graphs. In contrast, our framework dynamically constructs target-aware causal graphs and performs path-level validation and evidence aggregation during inference.

\subsection{Knowledge Graph-Augmented Reasoning}
Another line of work augments LLM reasoning with external knowledge graphs (KGs). Path-based methods select informative KG paths to support multi-hop reasoning~\cite{liu2024kelp,pog2025}. Recent benchmarks evaluate LLMs' use of KG structure~\cite{markowitz2025kgllmbench}, with causal-aware graph augmentation further explored~\cite{luo2025causal}. However, these methods treat paths primarily as evidence and do not explicitly model causal direction or competing effects. Our framework instead models signed causal influence and aggregates conflicting evidence during inference.


\section{Methodology}
\label{sec:method}

\textbf{Overview.} Given a context-free intervention-based causal question, the goal is to determine the directional effect of an intervention variable on a target outcome. To this end, we propose an explicit causal reasoning framework that constructs, audits, and aggregates causal evidence in a structured and interpretable manner. As illustrated in Fig.~\ref{fig:pipeline}, the framework consists of four sequential stages. First, structured query extraction identifies the intervention variable, target variables, and outcome modifiers from the input question. Second, a target-aware causal graph is constructed through iterative expansion, where graph growth is explicitly constrained by relevance to the target variable. Third, candidate causal chains connecting the intervention and target are extracted from the graph and audited to ensure causal consistency under hypothetical interventions. Finally, validated causal evidence from the first three stages is aggregated to produce the final decision on the direction of the intervention effect. This staged formulation enables systematic control of the reasoning process, improves interpretability through explicit causal structures, and allows only causally validated evidence to contribute to the final prediction.

\subsection{Causal Variable Identification }
\label{sec:problem_setup}

Given a natural language question $q$, we employ an LLM-based extractor $\Phi_{\text{extract}}$ to produce a structured query
\begin{equation}
\mathcal{Q} = (X, Y_s, Y_b, I).
\end{equation}
Here, $X$ denotes the \emph{intervention variable} representing the manipulated cause event; $Y_s$ is the \emph{surface target}, i.e., the surface-level outcome expression in the question; $Y_b$ is the \emph{base target variable}, a canonical scientific noun phrase used for causal reasoning; and $I$ contains \emph{outcome modifiers}, including the directional indicator $D \in \{\textsc{More}, \textsc{Less}\}$ and a negation flag. We distinguish $Y_s$ from $Y_b$ so that graph construction and causal reasoning are performed on the normalized variable pair $(X, Y_b)$, while the final prediction is mapped back to the original question semantics using $I$. This separation decouples causal reasoning from surface linguistic variation, enabling more stable graph construction and consistent reasoning across semantically equivalent questions.

\subsection{Target-Aware Causal Graph Construction}
\label{sec:graph_construction}

We propose to construct a directed causal graph $\mathcal{G}=(\mathcal{V}, \mathcal{E})$ rooted at
the intervention variable $X$, with the objective of identifying causal mechanisms
leading to the base target variable $Y_b$.

\paragraph{BFS Expansion with Negative Constraints.} We adopt a breadth-first search (BFS) expansion strategy~\cite{kozen1992depth} that iteratively explores candidate causal effects layer by layer to construct the graph. This process incrementally grows the graph outward from $X$, allowing the framework to capture multi-hop causal mechanisms while maintaining explicit control over the search depth and expansion scope. We initialize the frontier $\mathcal{F}_0=\{X\}$ and the visited set $\mathcal{V}_{\text{vis}}=\{X\}$, where the frontier represents the set of nodes to be expanded at the current step, and the visited set records all nodes observed
so far. At each expansion step $t$, we collect a pool of LLM-proposed candidate causal triples for the next layer, denoted by $\mathcal{C}_{t+1}$: 
\begin{equation}
\begin{aligned}
\mathcal{C}_{t+1}
&=
\bigcup_{u \in \mathcal{F}_t}
\Bigl\{ (u,r,v)\ \Bigm|\ (u,r,v)\sim \\
&\qquad \Phi_{\text{expand}}\!\left(u \mid \text{Avoid}=\mathcal{V}_{\text{vis}}\right)
\Bigr\}.
\end{aligned}
\end{equation}
Here, $(u,r,v)$ represents a directed causal relation $u\!\rightarrow\!v$ with effect polarity $r\in\{\textsc{Inc},\textsc{Dec}\}$, where \textsc{Inc} denotes $u\uparrow \Rightarrow v\uparrow$ and \textsc{Dec} denotes $u\uparrow \Rightarrow v\downarrow$. The $\sim$ denotes LLM generation or decoding.
Injecting the visited set as a negative constraint discourages revisiting previously explored nodes, reducing redundant loops and cyclic reasoning. This design enforces acyclic graph growth and prevents uncontrolled expansion, keeping the search focused on plausible causal mechanisms.

\paragraph{Fine-Grained Variable Alignment (FGVA).}
To determine whether a generated node $v$ is relevant to the base target $Y_b$, we introduce a fine-grained variable alignment function $\Phi_{\text{align}}(v, Y_b)$ that evaluates three dimensions: (i) \emph{Entity alignment} $S_E \in \{\text{Exact}, \text{Partial}, \text{None}\}$; (ii) \emph{Quantity alignment} $S_Q \in \{\text{Exact}, \text{Subset}, \text{Agg}, \text{None}\}$; and (iii) \emph{State alignment} $S_S \in \{\text{Match}, \text{Conflict}, \text{None}\}$. Intuitively, $S_E$ assesses whether the core entity or topic matches (or partially overlaps), $S_Q$ checks whether $v$ and $Y_b$ refer to the same granularity (e.g., exact vs. subset/aggregate), and $S_S$ checks whether the described state is logically compatible. By jointly considering these dimensions, FGVA enables relevance assessment beyond surface string similarity, allowing the framework to distinguish target-equivalent variables, near-target proxies, and intermediate bridging concepts.

Based on the alignment results, we classify each node $v$ into a relevance class:
\begin{equation}
\resizebox{0.8\linewidth}{!}{$
Cls(v)=
\begin{cases}
\textsc{Exact}, & S_E^{ex}\land S_Q^{ex}\land S_S^{m},\\
\textsc{CloseHit}, & S_E^{ex}\land S_Q^{sa}\land \neg S_S^{c},\\
\textsc{Bridge}, & S_E^{pa}\land \neg S_S^{c},\\
\textsc{None}, & \text{otherwise}.
\end{cases}
$}
\end{equation}
where $S_E^{ex}\!\triangleq\![S_E=\text{Exact}]$, $S_E^{pa}\!\triangleq\![S_E=\text{Partial}]$, $S_Q^{ex}\!\triangleq\![S_Q=\text{Exact}]$, $S_Q^{sa}\!\triangleq\![S_Q\in\{\text{Subset},\text{Agg}\}]$, $S_S^{m}\!\triangleq\![S_S=\text{Match}]$, and $S_S^{c}\!\triangleq\![S_S=\text{Conflict}]$. The assigned class $Cls(v)$ determines how the node is used during graph construction. Consequently, we maintain three auxiliary sets:
$\mathcal{N}_{\text{exact}}=\{v\mid Cls(v)=\textsc{Exact}\}$,
$\mathcal{N}_{\text{close}}=\{v\mid Cls(v)=\textsc{CloseHit}\}$, and
$\mathcal{N}_{\text{bridge}}=\{v\mid Cls(v)=\textsc{Bridge}\}$, while nodes with $Cls(v)=\textsc{None}$ are discarded and not expanded further. \textsc{Exact} nodes are treated as target-equivalent and serve as terminal endpoints for causal-chain extraction. \textsc{CloseHit} nodes act as semantic landing points near $Y_b$ and support subsequent alignment toward the target. \textsc{Bridge} nodes are retained only as intermediate candidates and may continue to be expanded to reach \textsc{CloseHit} or \textsc{Exact} nodes, helping reduce semantic drift while preserving recall.

\paragraph{Target-Aware Pruning and Frontier Update.}
To control the exponential growth of the search space during graph expansion, we introduce a target-aware pruning strategy at each BFS step. Unconstrained expansion rapidly introduces loosely related or generic nodes, diluting causal signals and increasing the risk of spurious reasoning paths~\cite{SUI2022108943}. Instead of heuristic string matching or static embedding similarity, we propose to use an LLM-based relevance ranking function $\Phi_{\text{rank}}(v, Y_b)$ that explicitly conditions on the base target variable $Y_b$. This ranking prioritizes nodes that are not only semantically related, but also causally promising for reaching the target domain. 

At each expansion step, only the top-$K$ ranked nodes are retained as the next frontier:
\begin{equation}
\mathcal{F}_{t+1} = \text{Top-}K(\mathcal{C}_{t+1}), ~~
\mathcal{V}_{vis} \leftarrow \mathcal{V}_{vis} \cup \mathcal{F}_{t+1}.
\end{equation}
This goal-directed pruning focuses the expansion on plausible causal mechanisms while maintaining coverage of relevant effects. The process terminates when $Y_b$ is reached or when a maximum depth $D$ is exceeded.

\paragraph{Bridging Close-Hit Nodes.}
When the BFS expansion does not explicitly reach $Y_b$, we perform a bridge step to connect near-target nodes in $\mathcal{N}_{close}$. For each $u \in \mathcal{N}_{close}$, a verifier $\Psi_{\text{bridge}}$ evaluates whether a direct causal relation is plausible; if so, a directed edge $(u, r_{\text{bridge}}, Y_b)$ is added to $\mathcal{E}$.

\subsection{Causal Chain Extraction and Auditing}
\label{sec:auditing}

To derive causally valid explanations for intervention effects, we extract and audit causal chains connecting the intervention variable $X$ to the base target $Y_b$ from the constructed graph $\mathcal{G}$. We enumerate simple paths $\mathcal{P} = \{p_1, \dots, p_m\}$ connecting $X$ and $Y_b$, rank them using a structural cost function that penalizes excessive length, semantic drift, and generic nodes, and retain the top-$K'$ paths for a two-stage causal audit.

\paragraph{Premise Consistency Check.}
A valid causal explanation must not contradict the intervention premise. For a path $p$, we define a premise validity indicator
\begin{equation}
V_{\text{prem}}(p)
=
\prod_{v \in p \setminus \{X, Y_b\}}
\mathbb{I}\!\left[\neg C(v, X)\right].
\end{equation}
where $C(v,X)$ indicates if introducing an intermediate node $v$ would violate the intervention premise on $X$. Paths with $V_{\text{prem}}(p)=0$ are discarded.

\paragraph{Counterfactual Edge Verification.}
Each remaining path is further audited via counterfactual probing. For each edge $e=(u,v)$, an LLM estimates whether removing or reversing $u$ induces a change in $v$, producing a edge-level confidence $V_{\text{cf}}(e) \in [0,1]$. The final path-level audit score is defined as
\begin{equation}
S_{\text{audit}}(p)
=
V_{\text{prem}}(p)
\cdot
\prod_{e \in p} V_{\text{cf}}(e).
\end{equation}
Paths with multiplicative audit score $S_{\text{audit}}(p)$ less than a predefined threshold $\tau_{\text{audit}}$ are removed.

\subsection{Evidence Aggregation and Final Decision}
\label{sec:decision}

\paragraph{Path-Level Effect and Weighting.}
Each causal relation $r$ is assigned a signed effect $\sigma(r)\in\{+1,-1\}$. The net causal effect of a path $p$ is computed as
\begin{equation}
\Delta(p)=\prod_{e\in p}\sigma(r_e),
\end{equation}
reflecting whether the intervention increases or decreases the target outcome. To downweight weakly supported or implicit reasoning, each path is assigned a weight
\begin{equation}
w(p)=S_{\text{audit}}(p)\cdot \exp(-\gamma N_{\text{br}}(p)),
\end{equation}
where $N_{\text{br}}(p)$ denotes the number of bridge edges. 

\paragraph{Graph-Level Evidence Aggregation.}
We aggregate signed evidence across all audited paths $\mathcal{P}^*$. Let $W^+$ and $W^-$ denote the total positive and negative evidence mass, respectively. The resulting graph-level confidence is defined as
\begin{equation}
C_{\text{graph}}
=
\frac{|W^+ - W^-|}{W^+ + W^- + \epsilon}
\cdot
\log\!\left(1 + |\mathcal{P}^*|\right),
\end{equation}
which jointly captures evidence dominance and support size. The corresponding graph decision $E_{\text{graph}}\in\{\textsc{More},\textsc{Less}\}$ is
determined by the sign of $W^+ - W^-$.

\paragraph{Conflict Resolution with Evidence-Conditioned LLM.} 
When causal evidence is weak or conflicting, we defer to an LLM prediction conditioned on the audited graph evidence. We serialize $\mathcal{P}^*$ into a compact evidence packet $\mathcal{Z}(\mathcal{P}^*)$, containing for each path its signed effect $\Delta(p)$, weight $w(p)$, and a short edge-sequence representation.\footnote{Deterministic decoding is used.} Formally,
\begin{equation}
E_{\text{LLM}} = f_{\theta}\!\big(q,\mathcal{Z}(\mathcal{P}^*)\big)\in\{\textsc{More},\textsc{Less}\}.
\end{equation}
The final prediction is obtained by resolving $E_{\text{graph}}$ and $E_{\text{LLM}}$ using the graph confidence:
\begin{equation}
\resizebox{0.85\linewidth}{!}{$
\hat{y} =
\begin{cases}
E_{\text{graph}}, & C_{\text{graph}} > \tau \ \text{or}\ E_{\text{graph}}=E_{\text{LLM}}, \\
E_{\text{LLM}}, & \text{otherwise}.
\end{cases}
$}
\end{equation}
Finally, $\hat{y}$ is mapped to the surface-level answer using the modifier set $I$.

\section{Experiments}
\label{sec:experiments}

\subsection{Datasets, Task, and Evaluation Metric}
\label{sec:datasets}
\textbf{Datasets and Task.} We evaluate our method on three benchmarks spanning diverse domains and knowledge requirements: (i) DDXPlus-CausalEffect (medical); (ii) the directional subset of WIQA; and (iii) a general-domain dataset derived from CauseNet. All datasets are cast into a unified \emph{context-free causal direction prediction} setting. Each instance specifies an intervention on a cause variable or event $X$ and asks whether the target variable or event $Y$ becomes \texttt{more} or \texttt{less} likely under the intervention. Following prior protocols on WIQA, we exclude NoEff instances and retain the original two-choice answer space. For DDXPlus-CausalEffect and the CauseNet-derived dataset, questions are rendered using a WIQA-style template with the same label set $\{\texttt{more},\texttt{less}\}$. This unification ensures consistent inference and fair comparison across datasets and methods. Dataset construction details and statistics are provided in Appendix~\ref{app:datasets}.

\noindent
\textbf{Evaluation Metric.} Following prior works, we report \emph{Accuracy} as the evaluation metric. Since the task is a binary directional classification with a fixed label space and follows standard evaluation protocols on the directional subset, Accuracy provides a clear and comparable measure across backbones and methods. Label distributions are reported in Appendix~\ref{app:datasets} for completeness.

\subsection{Experimental Setup}
\label{sec:impl}

\textbf{Decoding and Reproducibility.} All LLM calls use deterministic decoding (temperature $=0$) with fixed decoding hyperparameters (e.g., \texttt{top\_p} and \texttt{max\_tokens}). When supported by the backend, we additionally fix the random seed to improve run-to-run consistency. Outputs are strictly constrained via JSON/schema validation; invalid responses are retried using a predefined template fallback.

\noindent
\textbf{Prompt Format.}
All methods adopt structured prompts with a fixed output schema, where the final prediction is constrained to $\{\texttt{more},\texttt{less}\}$ and optional fields are used for intermediate reasoning or evidence. Full prompts, schemas, and example instances are provided in Appendix~\ref{app:prompts}.

\noindent
\textbf{Graph Construction and Auditing.}
Graph expansion is performed with bounded depth $D = 4$ and Top-$K = 2$ frontier selection per hop. For each frontier node, up to $R = 3$ candidate relations are generated, and simple paths from $X$ to the (base) target are enumerated up to length $L = 6$. Candidate paths are filtered using an audit threshold $\tau_{\text{audit}} = 0.6$ before evidence aggregation. All hyperparameters are selected on a held-out development split (or via the grid search in Sec.~\ref{sec:sensitivity}) and fixed across all test runs.

\subsection{Backbone Models and Compared Methods}
\label{sec:backbones}

\textbf{Backbone Models}
We evaluate our method on multiple instruction-tuned LLM backbones from different model families at comparable scales, including Llama-3.1-8B, Mistral-7B, and Ministral-3-8B. To isolate the effect of the proposed reasoning framework, we keep the dataset, instance representation (explicit $(X,Y)$), prompt format, input/output schema, and decoding hyperparameters fixed for each backbone, and vary only the backbone model. This allows us to assess whether performance gains generalize across model families
rather than relying on a specific backbone.

\noindent
\textbf{Compared Methods}
We compare our method with representative LLM-based reasoning baselines under a unified evaluation pipeline. Specifically, we include (i) \emph{Direct LLM}, which directly predicts the directional label from $(X,Y)$ without intermediate structure, serving as a parametric-knowledge baseline; (ii) \emph{Chain-of-Thought} (COT) prompting~\cite{wei2022chain}, which elicits free-form step-by-step rationales before prediction; (iii) structured reasoning methods such as \emph{Tree-of-Thought} (ToT)~\cite{yao2023tree} and \emph{Graph-of-Thought} (GoT)~\cite{besta2024graph}, which explore multiple reasoning paths via tree- or graph-structured deliberation without explicit causal modeling or verification; and (iv) CDCR-SFT~\cite{li2025mitigating}, which enhances causal reasoning by supervised LoRA fine-tuning on CausalDR, encouraging LLMs to explicitly construct causal DAGs before making predictions. All baselines use the same final answer constraint $\{\texttt{more},\texttt{less}\}$ and identical decoding settings unless otherwise specified.

\begin{table*}[t]
\centering
\renewcommand{\arraystretch}{1.1}
\resizebox{0.9\linewidth}{!} {
\begin{tabular}{llccccc}
\toprule
\textbf{Datasets} & \textbf{Backbone Models}
& \textbf{Direct LLM} & \textbf{CoT} & \textbf{GoT} & \textbf{ToT} & \textbf{Ours} \\
\toprule
\multirow{3}{*}{DDXPlus-CausalEffect}
& Llama-3.1-8B   & 66.00 & 59.50 & 57.50 & 58.00 & \textbf{67.00} \\
& Mistral-7B     & 43.50 & 40.00 & 46.00 & 38.00 &\textbf{74.50} \\
& Ministral-3:8b & 42.50 & 52.50 & 43.00 & 60.5 &\textbf{81.50} \\

\midrule
\multirow{3}{*}{WIQA (direction subset)}
& Llama-3.1-8B   & 56.13 & 50.00 & 51.42 & 50.94 &\textbf{67.92} \\
& Mistral-7B     & 31.13 & 37.74 & 41.51 & 37.26 &\textbf{65.09} \\
& Ministral-3:8b & 44.34 & 54.72 & 56.13 & 51.89 &\textbf{60.38} \\

\midrule
\multirow{3}{*}{CauseNet-derived (direction)}
& Llama-3.1-8B   & 68.00 & 77.00 & 73.00 & 77.00 &\textbf{79.00} \\
& Mistral-7B     & 72.00 & 81.00 & 67.00 & 50.00 &\textbf{87.00} \\
& Ministral-3:8b & 66.00 & 69.00 & 71.00 & 63.00 &\textbf{79.00} \\
\bottomrule
\end{tabular}
}
\vspace{-0.1cm}
\caption{Accuracy (\%) on context-free causal direction prediction across three datasets and multiple LLM backbones. Free-form reasoning methods (CoT, GoT, ToT) exhibit high variance across backbones. Ours consistently achieves the best performance across backbones and domains, demonstrating robust gains from explicit causal graph construction, auditing, and evidence aggregation. Accuracy is reported under a best-of-multiple-attempts setting.}
\label{tab:main_results}
\end{table*}

\begin{table}[t]
\centering
\renewcommand{\arraystretch}{1.1}
\resizebox{\linewidth}{!} {
\begin{tabular}{lccc}
\toprule
\textbf{Backbone Models} & \textbf{CDCR-SFT} & \textbf{Ours} & \textbf{$\Delta$} \\
\toprule
Llama-3.1-8B & 55.66 & \textbf{67.92} & +12.26 \\
Mistral-7B   & 44.81 & \textbf{65.09} & +20.28 \\
\bottomrule
\end{tabular}
}
\vspace{-0.1cm}
\caption{Accuracy (\%) on WIQA (direction subset) compared with CDCR-SFT, a training-based method that learns causal DAG construction via supervised fine-tuning. Our method achieves substantially higher accuracy on both backbones without task-specific fine-tuning, highlighting the effectiveness of explicit and auditable causal modeling at inference time.}
\vspace{-0.1cm}
\label{tab:wiqa_small_compare}
\end{table}

\subsection{Main Results}
\label{sec:main_results}

Table~\ref{tab:main_results} reports accuracy (\%) for context-free causal direction prediction across datasets and backbones. Overall, \textbf{ours} consistently outperforms all baselines. We highlight three recurring patterns.

\noindent
\textbf{(1) Reliability under domain-specific causal knowledge.} On DDXPlus-CausalEffect, which involves specialized medical knowledge and non-trivial causal dependencies, unconstrained reasoning baselines (CoT, ToT, GoT) are often unstable and can underperform Direct LLM prediction. This suggests that free-form rationales tend to introduce unsupported or domain-invalid mechanisms in expert domains. In contrast, \textbf{ours} consistently improves performance across all backbones, with particularly large gains on Mistral-7B (+28.5) and Ministral (+21.0). These results indicate that our proposed explicit construction and auditing of causal chains is crucial for reliable intervention reasoning in knowledge-intensive settings.

\noindent
\textbf{(2) Strong gains on complex multi-hop causal reasoning.}
On WIQA, which requires multi-hop causal reasoning under interventions, \textbf{ours} yields substantial improvements over the strongest baseline across all backbones (+11.8, +23.6, and +4.3 points for Llama-3.1-8B, Mistral-7B, and Ministral, respectively). Notably, existing reasoning paradigms exhibit high variance across backbones, whereas \textbf{ours} consistently achieves the best performance. This suggests that our proposed explicit construction and auditing of causal structures is more effective than relying on implicit deliberation alone, particularly for long-horizon causal inference. By converting global reasoning into target-aware expansion and local edge verification, our framework provides a reliable inductive bias that generalizes across model families.

\noindent
\textbf{(3) Robustness in open-domain causal reasoning.}
On the open-domain CauseNet-derived benchmark,  where spurious co-occurrence is common, purely deliberative or search-based reasoning methods exhibit high variance across
backbones. For example, ToT performs competitively on Llama-3.1-8B but collapses on Mistral-7B. In contrast, \textbf{ours} consistently achieves strong performance across all backbones, outperforming the best baseline on Mistral-7B (+6.0) and Ministral (+8.0). This robustness highlights the benefit of our proposed target-aware pruning and evidence aggregation in suppressing spurious causal paths and maintaining reliable inference in noisy, open-domain settings.

\noindent
\textbf{Comparison with causal-structure fine-tuning.}
To further contextualize our WIQA results, we compare against CDCR-SFT, a recent method that improves causal reasoning by supervised LoRA fine-tuning on the CausalDR dataset to explicitly learn causal DAG construction during training. As shown in Table~\ref{tab:wiqa_small_compare}, \textbf{ours} substantially outperforms CDCR-SFT on both backbones, with gains of +12.26 on Llama-3.1-8B and +20.28 on Mistral-7B. Notably, our approach achieves these improvements without any task-specific fine-tuning, demonstrating that explicit and auditable causal modeling at inference time can be more effective and generalizable than embedding causal structure implicitly into model parameters.

\begin{table}[t]
\centering
\small
\setlength{\tabcolsep}{4pt}
\begin{tabular}{c c c c c c}
\toprule
\multicolumn{3}{c}{\textbf{Hyperparameters}} & \multicolumn{3}{c}{\textbf{Performance}} \\
\midrule
\textsc{Depth} & \textsc{Width} & \textsc{PathLen} & Acc & Exo-Acc & In-Acc \\
($D$) & ($R$) & ($L$) & (\%) & (\%) & (\%) \\
\midrule
\multirow{4}{*}{2} 
 & 3 & 4 & 64.15 & 67.92 & 60.38 \\
 & 3 & 6 & 66.04 & 70.75 & 61.32 \\
 & 5 & 4 & 62.26 & 66.98 & 57.55 \\
 & 5 & 6 & 63.21 & 69.81 & 56.60 \\
\midrule
\multirow{4}{*}{4} 
 & 3 & 4 & 62.74 & 66.98 & 58.49 \\
 & 3 & 6 & \textbf{67.92} & 70.75 & \textbf{65.09} \\
 & 5 & 4 & 66.04 & 70.75 & 61.32 \\
 & 5 & 6 & 65.09 & 68.87 & 61.32 \\
\midrule
\multirow{4}{*}{6} 
 & 3 & 4 & 63.21 & 67.92 & 58.49 \\
 & 3 & 6 & 65.57 & 67.92 & 63.21 \\
 & 5 & 4 & 67.45 & \textbf{71.70} & 63.21 \\
 & 5 & 6 & 66.98 & 69.81 & 64.15 \\
\bottomrule
\end{tabular}
\vspace{-0.1cm}
\caption{Hyperparameter sensitivity analysis on WIQA (direction subset) with the Llama-3.1-8B backbone.}
\vspace{-0.1cm}
\label{tab:grid_search}
\end{table}

\subsection{Hyperparameter Sensitivity Analysis}
\label{sec:sensitivity}
We analyze the sensitivity of our target-aware causal graph construction to three key hyperparameters: maximum expansion depth ($D \in \{2,4,6\}$), branching width ($R \in \{3,5\}$), and maximum path length ($L \in \{4,6\}$). Results are reported in Table~\ref{tab:grid_search} on \textsc{WIQA} with the Llama-3.1-8B backbone.

\noindent
\textbf{Effect of Expansion Depth.}
Performance improves when increasing the depth from $D=2$ to a moderate range ($D=4$), reflecting the necessity of multi-hop causal mechanisms for WIQA-style intervention questions. However, further increasing depth ($D=6$) yields diminishing or unstable gains, suggesting that overly deep expansion introduces semantically weak or noisy intermediate variables. This trend supports our design choice of bounded expansion with auditing, which prioritizes causal relevance over exhaustive search.

\noindent
\textbf{Interaction Between Branching and Path Length.}
We observe a clear trade-off between branching width ($R$) and allowed path length ($L$). Moderate branching with sufficient path length (e.g., $R=3, L=6$) consistently yields strong performance, suggesting that accurate intervention reasoning relies on precise multi-hop causal chains rather than broad, shallow exploration. In contrast, aggressive early branching often degrades accuracy, particularly for shorter paths, due to the introduction of spurious or weakly related variables. These trends directly motivate our target-aware pruning and path-level auditing: by constraining early expansion and validating longer chains, our framework favors causally coherent multi-hop mechanisms while suppressing noisy alternatives.

\noindent
\textbf{Exogenous vs.\ Internal Interventions.} 
Exo-Acc and In-Acc measure accuracy when the intervention variable appears outside or inside the paragraph, respectively. Higher In-Acc reflects better alignment and reduced semantic drift, while gains on Exo-Acc indicate stronger context-free generalization. Exo-Acc benefits from larger expansion depth due to longer mechanism chains, whereas In-Acc typically peaks at moderate depth. Their complementary depth preferences validate our bounded, target-aware expansion design.

\subsection{Case Study and Interpretability}
\label{sec:case_study}

We present a representative case study to illustrate how explicit causal modeling improves both reliability and interpretability. While standard CoT reasoning produces free-form narratives whose causal validity is difficult to assess, our framework externalizes reasoning as explicit causal graphs, audited chains, and quantified evidence.
In a DDXPlus-CausalEffect example predicting the effect on \textit{Chagas probability}, CoT incorrectly outputs \texttt{more} by invoking a generic prior (``early-stage disease may be asymptomatic'') that is not causally grounded in the intervention. In contrast, our method constructs multiple candidate causal chains from the intervention variable, filters invalid mechanisms via counterfactual edge verification, and aggregates only premise-consistent evidence. This process yields the correct \texttt{less} prediction and exposes the specific causal paths supporting the decision. We provide the full constructed graph, audited chains, and per-edge counterfactual scores in Appendix~\ref{app:case_study}.

\section{Conclusion}
\label{sec:conclusion}

We propose an explicit, auditable causal reasoning framework for context-free intervention question answering. It formulates inference as structured reasoning via target-aware causal graph construction, path-level auditing, and evidence aggregation. By constraining variable expansion and verifying causal relations via counterfactual probing, it transparently resolves competing causal mechanisms. This design offers a clean paradigm for integrating LLMs with symbolic causal reasoning, transforming LLMs from end-to-end reasoners into constrained causal evaluators. Extensive experiments across medical, commonsense, and open-domain benchmarks demonstrate consistent and substantial improvements over existing methods, underscoring the importance of explicit causal structure and verification for reliable causal reasoning.

\section*{Limitations}
\label{sec:limitations}
\textbf{Inference Complexity}
The framework relies on multiple LLM interactions to support explicit causal graph construction and verification. Compared to single-pass approaches, this introduces additional inference steps, but the process is bounded by controlled expansion depth and pruning, reflecting a deliberate trade-off between reasoning reliability and efficiency.

\noindent
\textbf{Potential Risk of Misuse.} 
The framework produces directional causal predictions with structured reasoning traces, but does not perform formal causal identification based on do-calculus. In sensitive domains such as healthcare, the outputs should therefore be interpreted as supportive reasoning signals rather than definitive causal or clinical conclusions.

\bibliography{custom}

\newpage

\appendix

\twocolumn[
\begin{center}
{\Large\bfseries Appendix}
\vspace{2.5em}
\end{center}
]

\section{Algorithm Pseudocode}
\label{app:algo}

Algorithm~\ref{alg:framework} presents the pseudocode of the proposed framework.

\begin{algorithm}[htbp]
\small
\caption{Target-Aware Causal Graph Reasoning}
\label{alg:framework}
\SetKwInOut{Input}{Input}
\SetKwInOut{Output}{Output}

\Input{Question $q$}
\Output{Predicted answer $\hat{y}$}

\tcp{1. Setup (Sec. \ref{sec:problem_setup})}
$(X, Y_s, Y_b, I) \leftarrow \Phi_{\text{extract}}(q)$\;
$\mathcal{F} \leftarrow \{X\}$; \quad $\mathcal{V}_{vis} \leftarrow \{X\}$; \quad $\mathcal{N}_{close} \leftarrow \emptyset$\;

\tcp{2. Graph Expansion (Sec. \ref{sec:graph_construction})}
\While{$Y_b \notin \mathcal{F}$ \textbf{and} depth $< D$}{
    \tcp{Generate candidates avoiding history}
    $\mathcal{C} \leftarrow \Phi_{\text{expand}}(\mathcal{F} \mid \text{Avoid}=\mathcal{V}_{vis})$\;
    
    \tcp{Identify Close Hits via FGVA}
    \ForEach{$v \in \mathcal{C}$}{
        \If{$\Phi_{\text{align}}(v, Y_b) = \textsc{CloseHit}$}{
            $\mathcal{N}_{close} \leftarrow \mathcal{N}_{close} \cup \{v\}$\;
        }
    }
    
    \tcp{Prune to Top-K}
    $\mathcal{F} \leftarrow \Phi_{\text{rank}}(\mathcal{C}, Y_b, K)$\;
    $\mathcal{V}_{vis} \leftarrow \mathcal{V}_{vis} \cup \mathcal{F}$\;
    Update Graph $\mathcal{G}$\;
}

\tcp{3. Bridging (Sec. \ref{sec:graph_construction})}
\ForEach{$u \in \mathcal{N}_{close}$}{
    \If{$\Psi_{\text{bridge}}(u, Y_b)$ is valid}{
        Add edge $(u \to Y_b)$ to $\mathcal{G}$\;
    }
}

\tcp{4. Extraction \& Auditing (Sec. \ref{sec:auditing})}
$\mathcal{P}^* \leftarrow \text{Top-kPaths}(\mathcal{G}, X \to Y_b)$\;
\ForEach{$p \in \mathcal{P}^*$}{
    \tcp{Check Premise \& Counterfactuals}
    \If{$V_{\text{prem}}(p) = 0$ \textbf{or} $V_{\text{cf}}(p) < \tau$}{
        Discard path $p$\;
    }
}

\tcp{5. Decision (Sec. \ref{sec:decision})}
$E_{\text{graph}} \leftarrow \text{AggregatePaths}(\mathcal{P}^*)$\;
$E_{\text{LLM}} \leftarrow \Phi_{\text{reason}}(q, \mathcal{P}^*)$\;
$\hat{y} \leftarrow \text{ResolveConflict}(E_{\text{graph}}, E_{\text{LLM}})$\;

\Return $\hat{y}$ mapped to options via $I$\;
\end{algorithm}

\section{Dataset Construction Details}
\label{app:datasets}

\subsection{Overview}
Most existing causal question answering resources are \emph{context-dependent}, where each
question is paired with a passage and models are evaluated on evidence extraction
\cite{jin2023cladder,bondarenko2022causalqa}. In contrast, we target
\emph{context-free causal direction prediction} with \emph{graph-grounded explanations}:
each instance is defined over explicit variables/events $(X,Y)$ and the system predicts whether
intervening on $X$ makes $Y$ \emph{more} or \emph{less} likely.

We focus on the directional subset with labels $\{\texttt{more},\texttt{less}\}$.
Some datasets include \texttt{no effect} as a distractor option, but we exclude \texttt{NoEff}
instances as gold labels to keep the evaluation protocol consistent.
Table~\ref{tab:app_dataset_stats} summarizes dataset statistics reports how often questions include \emph{surface modifiers} (e.g., ``more/less probability of $Y$'').

\begin{table}[t]
\centering
\small
\setlength{\tabcolsep}{6pt}
\begin{tabular}{lccc}
\toprule
\textbf{Statistic} & \textsc{DDXPlus} & \textsc{WIQA} & \textsc{CauseNet} \\
\midrule
Domain          & Medical     & Science     & Open-domain \\
$N$             & 200         & 212         & 100 \\
More            & 100         & 99          & 50 \\
Less            & 100         & 113         & 50 \\
Avg.\ words     & 23.7        & 15.8        & 9.7 \\
Neg.\ (\%)      & 47.5        & 10.4        & 0.0 \\
\bottomrule
\end{tabular}
\caption{\textbf{Dataset statistics.} Directional subset only (\texttt{more}/\texttt{less}).
Negation via keyword match (e.g., \emph{not/no/without}) on the question stem.}
\label{tab:app_dataset_stats}
\end{table}

\subsection{DDXPlus-CausalEffect: Directional Effect Prediction from Medical Records}
We use the official DDXPlus release \cite{fansi2022ddxplus}, including condition metadata, evidence metadata,
and patient-level predefined splits. Each instance specifies an intervention variable $X$ (a patient evidence)
and an outcome variable $Y$ (a pathology), and asks how changing $X$ affects the probability of $Y$.

\paragraph{Association-based labeling (Stats Mode).}
Our primary construction assigns labels using empirical association statistics computed from patient records.
For each pathology $Y$ and evidence $X$, we compute $n_Y$, $n_X$, $n_{XY}$, and $N$ (total patients),
and estimate:
\begin{equation}
p(Y \mid X) = \frac{n_{XY}}{n_X}, \qquad
p(Y \mid \neg X) = \frac{n_Y - n_{XY}}{N - n_X},
\end{equation}
then define $\Delta = p(Y\mid X)-p(Y\mid \neg X)$.
Using a margin threshold $\tau$, we assign:
\begin{equation}
\texttt{label}(X,Y)=
\begin{cases}
\texttt{more}, & \Delta > \tau \\
\texttt{less}, & \Delta < -\tau \\
\texttt{discard}, & |\Delta| \le \tau.
\end{cases}
\end{equation}
These labels are treated as \textbf{data-driven directional proxies} (association-based) rather than fully
identified causal effects; discarding near-zero cases avoids ambiguous instances when statistics are insufficient.

\paragraph{Question rendering.}
We render each pair $(X,Y)$ into a WIQA-style question with answer choices restricted to
$\{\texttt{more},\texttt{less}\}$ (optionally keeping \texttt{no effect} only as a distractor).
An optional LLM rewriting step can improve fluency while preserving explicit $X$ and $Y$.

\subsection{WIQA Directional Subset}
WIQA \cite{tandon2019wiqa} is the closest widely-used benchmark aligned with our evaluation format because it
is framed as effect-direction prediction and its questions are formulated over explicit standalone events.
Following prior work, we evaluate on a curated subset that contains only directional-effect instances
(\texttt{more}/\texttt{less}) and excludes \texttt{NoEff} cases.

\subsection{CauseNet-derived Context-free Directional QA}
We construct a general-domain dataset from CauseNet \cite{heindorf2020causenet}.
Each record provides a head--tail causal assertion with optional confidence-like fields. We (i) sanitize concepts,
(ii) normalize heterogeneous confidence signals to a unified edge confidence $c(e)\in[0,1]$,
and (iii) map relations to signed directed edges with $\mathrm{sign}\in\{+1,-1\}$
(defaulting to $+1$ if polarity is missing).

\paragraph{Multi-hop path sampling and labeling.}
We sample simple directed paths $\pi=(x_0\!\rightarrow\!\cdots\!\rightarrow\!x_h)$
and derive the gold direction label by polarity product:
\begin{equation}
\texttt{label}(\pi)=
\begin{cases}
\texttt{more}, & \prod_{e\in\pi}\mathrm{sign}(e)=+1,\\
\texttt{less}, & \prod_{e\in\pi}\mathrm{sign}(e)=-1.
\end{cases}
\end{equation}
We keep only directional instances to align with our evaluation protocol.

\paragraph{Question rendering.}
For each sampled path, we form a WIQA-style context-free question over endpoints $X=x_0$ and $Y=x_h$
and store the underlying sampled path as an auditable explanation trace.

\paragraph{Limitations.}
This construction inherits noise from open IE extraction and may underrepresent negative relations when
explicit inhibitory edges are sparse; we therefore report label distributions and key sampling hyperparameters
alongside results.

\section{Prompt Templates and Output Constraints}
\label{app:prompts}

\subsection{Common I/O Format}
All datasets are cast into a unified \textbf{context-free directional-effect} format:
each instance specifies explicit variables/events $(X,Y)$ and asks whether intervening on $X$
makes $Y$ \textbf{more} or \textbf{less} likely. Following prior WIQA protocols, we exclude
\textsc{NoEff} and keep a \textbf{two-choice} setting. Accordingly, all prompts in this paper
share the same answer space: \textbf{A: more}, \textbf{B: less}.
All LLM calls use deterministic decoding with \texttt{temperature}=0.

\subsection{Answer Extraction and Fallback}
For all methods, we enforce an explicit final answer line:
\texttt{Final answer: A} or \texttt{Final answer: B}.
We extract the choice using a regex matcher. If the required line is missing, we apply a lightweight
\emph{forced extractor} prompt that maps the model output to $\{A,B\}$; if extraction still fails,
we fall back to \texttt{A}.

\subsection{Baseline Prompts}
\paragraph{Direct LLM.}
The model receives only the question stem and the two options, and must output the final choice:
\begin{quote}\small\ttfamily
[Direct]\;
\{question\_stem\}\\
A) more \quad B) less\\
Output format: Final answer: <A|B>
\end{quote}

\paragraph{Chain-of-Thought (CoT).}
We elicit a brief rationale before producing the final choice, while keeping the same answer constraint:
\begin{quote}\small\ttfamily
[CoT] Write brief causal reasoning, then choose A/B.\\
Output format:\\
Reasoning: <1--4 sentences>\\
Final answer: <A|B>
\end{quote}

\paragraph{Graph-of-Thought (GoT).}
GoT follows a split $\rightarrow$ analyze $\rightarrow$ merge procedure with explicit intermediate artifacts.
We first request a JSON decomposition into three components; then we analyze each component and produce
a merged solution. All stages are deterministic and the final response must end with the forced choice:
\begin{quote}\small\ttfamily
Split: Output ONLY JSON \{Component 1,2,3\}.\\
Analyze each component.\\
Merge and MUST end with: Final answer: <A|B>.
\end{quote}

\paragraph{Tree-of-Thought (ToT).}
ToT performs a deterministic tree-style deliberation with explicit candidate generation and selection:
(i) generate two candidate reasoning structures; (ii) select the better one;
(iii) generate two candidate reasoning traces conditioned on the selected structure;
(iv) select the better trace; (v) output the final choice.
The final output is forced into $\{A,B\}$:
\begin{quote}\small\ttfamily
Generate 2 candidates (JSON) $\rightarrow$ Pick 1/2\\
Generate 2 traces (JSON) $\rightarrow$ Pick 1/2\\
Final: MUST end with: Final answer: <A|B>
\end{quote}
We refer to these as GoT/ToT because they enforce graph-/tree-structured intermediate representations and
selection steps, even though decoding is deterministic in our implementation.

\subsection{Our Framework: Graph-grounded + Audit Prompts}
Our method uses the LLM as a \emph{structured evidence worker} and a final \emph{evidence aggregator}.
All intermediate prompts are constrained to JSON outputs to ensure robust parsing, while the final decision
is mapped to the unified two-choice space (A/B).

\paragraph{P1: Variable extraction (strict JSON).}
\begin{verbatim}
You are a STRICT string-matching text 
extractor.
Your job is ONLY to extract spans from 
the given question text.
Do NOT paraphrase. Copy spans exactly.

Question:
"{question}"

Output ONLY valid JSON (no extra text):
{
  "cause_event": "<copied text>",
  "outcome_text_raw": "<copied text>",
  "outcome_direction": "MORE" or "LESS" 
  or "NONE",
  "is_negated": true or false
}
\end{verbatim}

\paragraph{P2: Target-aware single-hop expansion (forward).}
\begin{verbatim}
You are a causal edge finder.

Input:
- CAUSE_NODE (X): "{X}"
- TARGET_HINT (Y): "{target_hint}"
- FORBIDDEN LIST (Avoid revisits): 
[{avoid_str}]

Task:
- Propose up to {max_relations} 
SINGLE-HOP causal effects 
starting from X.
- If TARGET_HINT is not "NONE", 
expand toward Y (prefer intermediates
that connect X to Y).
- Each tail must be a NEUTRAL NOUN
PHRASE (no "more/less", no full 
sentences).
- Use "INCREASES" when increasing 
head tends to increase tail.
- Use "DECREASES" when increasing
head tends to decrease tail.

Output ONLY JSON:
{
  "triples": [
    ["{X}", "INCREASES" | "DECREASES", 
    "<neutral noun phrase>"]
  ]
}
\end{verbatim}

\paragraph{P3: Target equivalence / bridging judgment.}
\begin{verbatim}
You are judging the relationship 
between two variables in a causal system.

Variable A: "{A}"
Variable B: "{B}"

Task:
Decide their relationship along 
three axes:
1) core_entity_relation
2) quantity_relation
3) causal_or_structural_relation

Output ONLY JSON:
{
  "core_entity_relation": "...",
  "quantity_relation": "...",
  "causal_or_structural_relation": 
  "...",
  "explanation": "short explanation"
}
\end{verbatim}

\paragraph{P4: Counterfactual edge audit (binary validity).}
\begin{verbatim}
You are a Scientific Logic Judge.

We MUST judge causality based on 
intervention semantics:
If we actively increase A, 
does B tend to increase/decrease?

Candidate causal edge:
A: "{A}"
Relation: "{REL}"   (INCREASES/DECREASES)
B: "{B}"

Return a conservative judgment. If unsure,
mark false.

Output ONLY JSON:
{"is_valid_link": true/false, "reasoning":
"short explanation"}
\end{verbatim}

\paragraph{P5: Final aggregation with audited chains.}
\begin{verbatim}
You are solving a WIQA-style causal
reasoning problem.
Your job is to decide how the CAUSE affects
the BASE VARIABLE.

Question: "{question}"
Cause event (X): "{cause_event}"
Outcome event (surface): "{outcome_event}"
BASE variable (outcome_base, 
the only quantity you judge): "{outcome_base}"

Causal graph summary:
{summary_json}

Evidence chains from cause → 
BASE (system-computed net effects; 
DO NOT re-multiply signs yourself):
{evidence_block}

IMPORTANT:
- Decide the direction of change for 
the BASE VARIABLE only ("{outcome_base}").
- [Net Effect: POSITIVE (Causes Increase)] 
supports "more" (A).
- [Net Effect: NEGATIVE (Causes Decrease)] 
supports "less" (B).
- If chains conflict, prefer higher-quality 
/ fewer-bridge chains.

Output ONLY strict JSON:
{
  "effect_on_base": "more" | "less",
  "final_answer": "A" | "B",
  "confidence": "high" | "medium" 
  | "low" | "very_low",
  "reasoning": "short explanation grounded in
  the evidence chains"
}
\end{verbatim}

\paragraph{Example evidence\_block (serialized chains).}
\begin{verbatim}
- Chain: X -> INCREASES -> M1 -> 
DECREASES -> outcome_base
  [Net Effect: NEGATIVE (Causes Decrease)]
  [bridge_edges: 0/2]
- Chain: X -> INCREASES -> M2 -> INCREASES
-> outcome_base
  [Net Effect: POSITIVE (Causes Increase)]
  [bridge_edges: 1/2]
\end{verbatim}

\section{Extended Case Study}
\label{sec:appendix}
\label{app:case_study}

\paragraph{Instance.}
Intervention:
$X=$\textit{the patient has not noticed any new fatigue, vague discomfort, diffuse
muscle aches, or a change in well-being}.
Target:
$Y_b=$\textit{Chagas probability}.
Gold label: \texttt{less}.

\paragraph{Baseline (CoT) output.}
CoT predicts \texttt{more} by invoking a generic narrative that Chagas can be
asymptomatic in early stages, treating symptom absence as evidence for infection.

\paragraph{Representative generated causal triples.}
Our graph construction produces localized, inspectable hypotheses as directed triples.
Examples include:
\begin{itemize}
  \item $X \xrightarrow{\textsc{Dec}}$ \texttt{patient's overall health status}
  \item $X \xrightarrow{\textsc{Inc}}$ \texttt{Chagas disease risk factors}
  \item \texttt{Chagas disease risk factors} $\xrightarrow{\textsc{Inc}}$ \texttt{infection probability}
  \item \texttt{infection probability} $\xrightarrow{\textsc{Dec}}$ \texttt{Chagas probability}
  \item \texttt{Chagas disease risk factors} $\xrightarrow{\textsc{Inc}}$ \texttt{Chagas probability}
  \item \texttt{patient's overall health status} $\xrightarrow{\textsc{Inc}}$ \texttt{Chagas probability}
\end{itemize}

\paragraph{Extracted causal chains.}
From the constructed graph, our path extractor returns three chains:
\[
\begin{array}{ll}
\text{(P1)} & X \xrightarrow{\textsc{Dec}} \texttt{overall health status}
              \xrightarrow{\textsc{Inc}} Y_b, \\
\text{(P2)} & X \xrightarrow{\textsc{Inc}} \texttt{risk factors}
              \xrightarrow{\textsc{Inc}} Y_b, \\
\text{(P3)}\;& X \xrightarrow{\textsc{Inc}} \texttt{risk factors} \\ 
            & \xrightarrow{\textsc{Inc}} \texttt{infection probability} \xrightarrow{\textsc{Dec}} Y_b.
\end{array}
\]

\paragraph{Counterfactual edge audit.}
For each edge $e=(u,v)$ on a candidate path, we compute a counterfactual support score
$V_{\text{cf}}(e)\in[0,1]$ via intervention-style probing, and define the path audit score as
\begin{equation}
S_{\text{audit}}(p)=V_{\text{prem}}(p)\cdot \prod_{e\in p} V_{\text{cf}}(e).
\end{equation}
In this example, the retained edges receive consistently high counterfactual support
(approximately $0.9$--$1.0$), and the final decision is made by aggregating the audited paths.

\paragraph{Aggregation and decision.}
The audited chains contain directional disagreement (one positive vs.\ two negative).
Our conflict-aware aggregation weights each chain by its audit score and combines
positive/negative evidence mass, resulting in a net \texttt{less} effect on $Y_b$ and a
final confidence of 0.85, matching the gold label.

\section{Ethics Statement}
This manuscript is the authors’ original work. Except for minor English grammar checking with ChatGPT, no large language model or AI tool was used for idea generation, problem formulation, literature search or screening, methodology design, code implementation, data processing, experimental design, statistical analysis, figure or table drafting, or substantive writing. All intellectual contributions, including conceptualization, model design, and empirical evaluation, are solely those of the authors.

\end{document}